\documentclass[conference]{IEEEtran}
\usepackage{cite}

\usepackage[pdftex]{graphicx}
\graphicspath{./figures/}

\usepackage{url}

\begin{document}

\title{Cognitive Architecture for Mutual Modelling}

\author{\IEEEauthorblockN{Alexis Jacq$^{1,2}$, Wafa Johal$^1$, Pierre Dillenbourg$^1$, Ana Paiva$^2$}
\IEEEauthorblockA{
$^1$CHILI Lab, \'Ecole Polytechnique F\'ed\'erale de Lausanne (EPFL), Lausanne,
Switzerland}
\IEEEauthorblockA{
$^2$INESC-ID \& Instituto Superior T\'{e}cnico, University of Lisbon, Portugal
}
}

\maketitle

\begin{abstract}
In social robotics, robots needs to be able to be understood by humans.
Especially in collaborative tasks where they have to share mutual knowledge. 
For instance, in an educative scenario, learners share their knowledge and they must adapt their behaviour in order to make sure they are understood by others. 
Learners display behaviours in order to show their understanding and teachers adapt in order to make sure that the learners' knowledge is the required one. 
This ability requires a model of their own mental states perceived by others: \textit{``has the human understood that I(robot) need this object for the task or should I explain it once again ?"}  
In this paper, we discuss the importance of a cognitive architecture enabling second-order Mutual Modelling for Human-Robot Interaction in educative contexts.

\end{abstract}


%
\IEEEpeerreviewmaketitle

\section{Introduction}
%
%
%
%
%
%
%
%


A social robot is brought to interact with humans. The quality of this interaction depends on its ability to behave in an acceptable and understandable manner by the user. Hence the importance for a robot to take care of his image: how much it is perceived as an automatic and repetitive agent, or contrarily as a surprising and intelligent character. If the robot is able to detect this perception of itself, it can adapt its behaviour in order to be understood: ``you think I am sad while I am happy, I want you to understand that I am happy". 

In a collaborative context, where knowledge must be shared, agents must exhibit that they are acquiring the shared information with an immediate behaviour: ``I look at what you are showing me, do you see that I am looking at it, do you think I am paying attention to your explanation ?"; ``I have understood your idea, do you understand that I have understood ?". 
As humans, we have different strategies to exhibit understanding or to resolve a misunderstanding. 
As an example, if someone is talking about a visual object, we alternatively gaze between the object and the person to make sure he saw that we gazed at the object. Or if we detect that the other person has not understood a gesture (e.g. pointing at an object) we would probably exaggerate the gesture.

Developed by Baron-Cohen and Leslie~\cite{baron1985does}, the Theory of Mind (ToM) describes the ability to attribute mental states and knowledge to others. In interaction, humans are permanently collecting and analysing huge quantity of information to stay aware of emotions, goals and understandings of their fellows. In this work, we focus on a generalization of this notion: Mutual Modelling characterizes the effort of one agent to model the mental state of another one~\cite{dillenbourg1999you}. 

Until now, the work conduced by the Human-Robot Interaction (HRI) community to develop mutual modelling abilities in robots was limited to a first level of modelling (see related work in section~\ref{rw}). Higher levels require the ability to recursively attribute a theory of mind to other agents (\textit{I think that you think that} ...) and their application to HRI remains unexplored. However, a knowledge of oneself perceived by others is necessary to adapt a behaviour to keep mutual understanding. 


An important challenge of social robotics is to provide assistance in education. 
The ability of robots to support adaptive and repetitive tasks can be valuable in a learning interaction.
The CoWriter Project~\cite{Hood,jacq2016building} introduces a new approach to help children with difficulties in learning handwriting. 
Based on the \emph{learning by teaching} paradigm, the goal of the project is not only to help children with their handwriting, but mainly to improve their self-confidence and motivation in practising such exercise.

\emph{Learning by teaching} engages students to conduct the activity in the role of the teachers in order to support their learning process. 
This  paradigm is known to produce motivational, meta-cognitive and educational benefits in a range of disciplines~\cite{Rohrbeck2003}. 
The CoWriter project is the first application of the learning by teaching approach to handwriting.

The effectiveness of this learning by teaching activity is built on the ``prot\'eg\'e effect'': the teacher feels responsible for his student, commits to the student's success and possibly experiences student's failure as his own failure to teach. 
The main idea is to promote the child's extrinsic motivation to write letters (he does it in order to help his ``prot\'eg\'e" robot) and to reinforce the self-esteem of the child (he plays the teacher and the robot actually progresses).

In that context, the robot needs to pretend enough difficulties to motivate the child to help it. 
This ability of the robot to pretend strongly depends on the perception of the robot by the child: the displayed behaviours (gestures, gazes and sounds) by the robot, the initial level and learning speed of the robot must match with what the child imagines of a ``robot in difficulty".
In order to adapt to the child, the robot needs then to have a model of how it is perceived by the child. On the other side, the child builds also a model of the robot's difficulties and attitude. 
This mutual-modelling is primordial in order to have mutual understanding and fluid interaction between learner and teacher. 

\section{Related works}\label{rw}

A large amount of fields have introduced frameworks to describe mutual modelling ability~\cite{lemaignan2015mutual}. 
In developmental psychology Flavell~\cite{flavell1990developmental} denotes two different levels of perspective taking: the \textit{cognitive connection} (I see, I hear, I want, I like...) and \textit{mental representation} (what other agents feel, hear, want...).

From a computational perspective, Epistemic logic describes knowledges and beliefs shared by agents. This framework enables consideration of infinite-level of mutual modelling. It defines a \textit{shared-knowledge} (all the agents of a group know \textbf{X}) and a \textit{common-knowledge} (all the agents of a group know \textbf{X}, and know that all the agent know \textbf{X}, and know that all the agents know that all the agents know \textbf{X}, \dots.)~\cite{hendricks2008epistemic}. 

Mutual modelling has also been studied through educational contexts. Roschelle and Teasley~\cite{roschelle1995construction} suggested that collaborative learning requires a \textit{shared understanding} of the task and of the shared information to solve it. 
The term ``mutual modelling" was introduced in Computer-Supported Collaborative Learning (CSCL) by Dillenbourg~\cite{dillenbourg1999you}. It focused on knowledge states of agents. Dillenbourg developed in \cite{sangin2007partner} a computational framework to represent mutual modelling situations.

However, HRI research has not, until now, explored the whole potential of mutual modelling. In \cite{scassellati2002theory}, Scassellati supported the importance of Leslie's and Baron-Cohen's theory of mind to be implemented as an ability for robots. 
He focused his work on attention and perceptual processes (face detection or colour saliency detection). Thereafter, some works (including Breazeal~\cite{breazeal2006using}, Trafton~\cite{Trafton2005}, Ros~\cite{Ros2010} and Lemaignan~\cite{lemaignan2012thesis}) were conduced to implement Flavell's first level of perspective taking~\cite{flavell1977development} (``\textit{I see (you do not see the book)}"), ability that is still limited to visual perception. 

Breazeal~\cite{breazeal2009embodied} and Warnier~\cite{warnier2012when} reproduced the Sally and Anne's test of Wimmer~\cite{wimmer1983beliefs} with robots able to perform visual perspective taking. The robot was able to infer the knowledge of a human given the history of his visual experience. 

In \cite{lemaignan2016realtime}, Lemaignan implemented a system that computes the visual field of agents and estimates which objects are looked at in real-time. This time, the robot is not just aware of what \textit{can be seen} by agents, but it perceives what \textit{is currently being looked at}. Lemaignan used this system to measure Sharma's \textit{with-me-ness}~\cite{sharma2014me}, visual commitment based on expected focus of attention in an activity. 

\section{MM-based reasoning}
A first intuition for mutual modelling is to assume that all agents have the same basic architecture. In~\cite{breazeal2006using}, Breazeal show a MM-based reasoning where the robot uses its own architecture to model other agents. We can imagine a second level of modelling where the robot recursively attribute to other agents the mutual modelling ability. But it would create an infinite recursive loop: the agent then models the robot that models the agent etc. Another reason to avoid an infinite recursive approach is that different agents can have different behaviours: in similar situations, they do not necessarily take similar decisions. 

We propose a different approach of modelling, where we define two orders of agents: the \textit{first-order-agents} deal with direct representations of agents by the robot (for example the child), while the \textit{second-order-agents} deal with the representation of agents by agents (for example the \textit{robot-perceived-by-the-child}). 
Modelling \textit{second-order-agent} like the \textit{robot-perceived-by-the-child} will help to model how the child perceives the robot, e.g. to make sure the child understands that the robot is learning from his demonstrations. 
We can also define \textit{$n^{\textit{th}}$-order} agents with a higher level of theory of mind. But taking into account high levels of mutual modelling would be difficult to process in real time. Unlike the epistemic logic, our proposed framework will not take into account infinite regress~\cite{clark1991grounding} of mutual modelling.

All sensors (cameras, micros, motor positions etc. and in the case of CoWriter the tablet's inputs) are used to perceive information about the physical behaviour of agents. 
We call all the measurable quantities or qualities that provide information like position in space, the direction of the gaze, speech, movement and facial expressions etc. as \textit{perceived variables}. 
Each agent's model is associated with a set of perceived variables that describes his physical behaviour.

Emotional states of agents cannot be directly measured directly from sensors. 
We call \textit{abstract variables} all the quantities or qualities that describe the mental state of an agent. 
Abstract variables are deduced from the dynamic of perceived variables. 
As an example, if the robot points at an object with its arm, it expects the child to look at the object. 
If then the child looks at the hand of the robot, the robot can deduce that the child has not understood the meaning of its gesture. 
The perceived variables are the robot's gesture and the gaze direction of the child. 
The deduced abstract variable is the understanding of the gesture by the child. 

A model of an agent is the set of all the values of the perceived or abstract variables associated with this model. 
Since the values of variables are likely to change with the time, the models must be dynamic.


In order to deduce the values of abstract variables (that can't be obtained from direct perception), we propose to build a Bayesian model based on the knowledge from perceived variables. 
The choice of a probabilist approach instead of a symbolic approach comes from the errors in the perception of the robot: knowledge and other mental states of agents can not be directly perceived through the behaviours. 
They must be inferred, hence a probabilistic model enables richer predictions.

This Bayesian network would contain the probabilities that \emph{abstract variables} take values given the values of \emph{perceived variables}. 

For example, if the robot points at an object and detects that the child saw the movement, then it expects the child to look at the targeted object immediately after.
In other terms, if the child looks at the hand of the robot but does not look at the target object, the probability that the child understood the pointing movement is expected to be small.
Knowing that, the robot can make the decision to exaggerate its pointing gesture.

\section{Description of the architecture}
\begin{figure}[!]
\centering
\includegraphics[width=1\columnwidth]{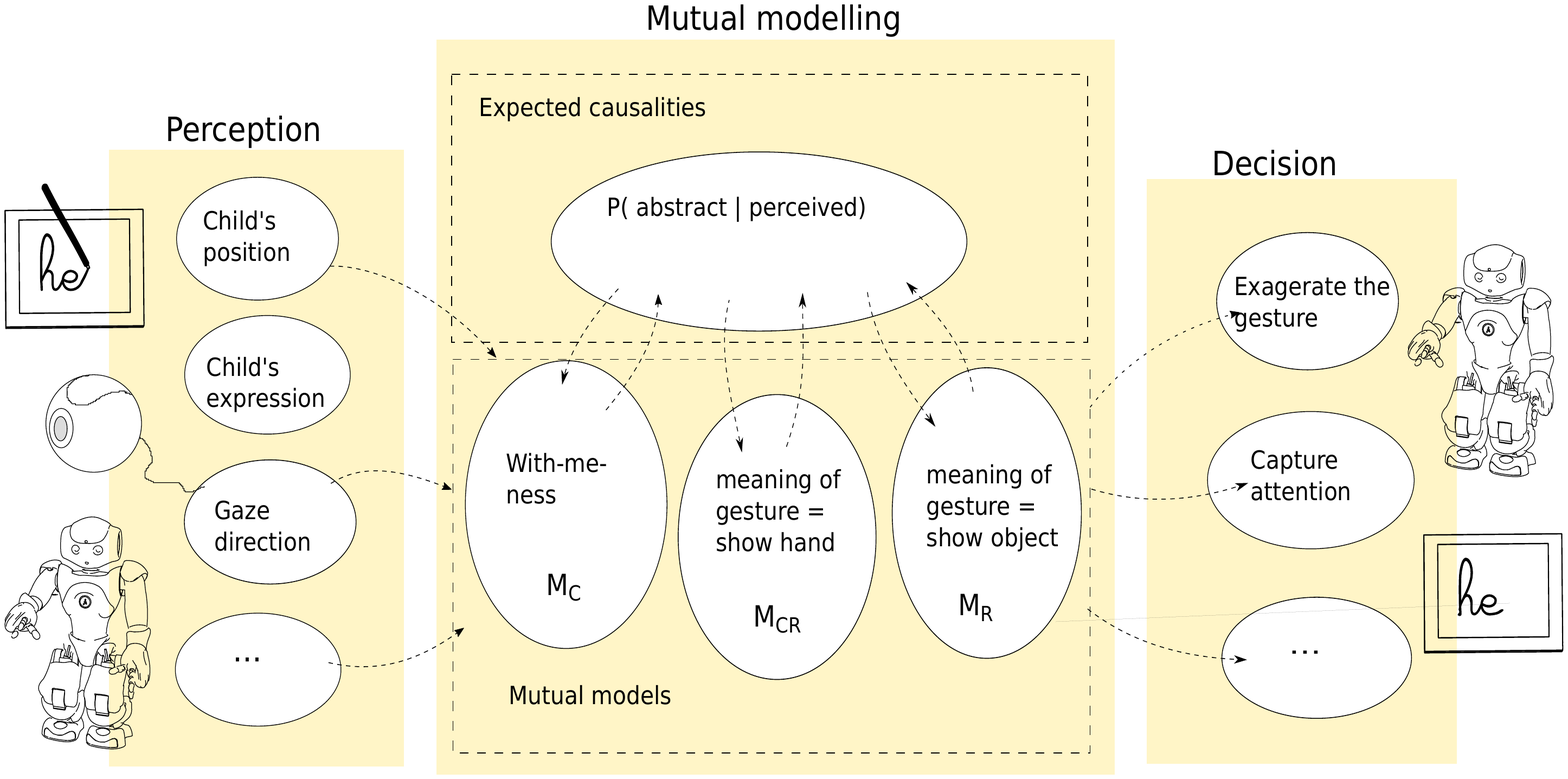}
\caption{\small\textbf{Overview of the cognitive architecture}. Yellow squares represent the main parts.
White ellipses represent modules. 
It shows possible devices used for perception and decisions with the context of the CoWriter Activity. 
We illustrate the architecture with a situation of misunderstanding : the child has a bad interpretation of the gesture of the robot. 
In order to resolve this misunderstanding, a possible decision could be to exaggerate the movement. }
\label{cog}
\end{figure}

The picture~\ref{cog} visually summarizes the global design of our architecture. 

Our cognitive architecture for mutual modelling contains three main parts. 
The \textbf{perception part} (see \ref{ssec:perception}) regroups all the modules that measure the values of the perceived variables using sensors. 
These values are sent to the \textbf{mutual modelling part} (see \ref{ssec:mmm}) that updates models with measured values of perceived variables and infers the value of abstract variables in real-time. 
Finally, the \textbf{decision part} (see \ref{ssec:decision}) contains all the modules associated to the control of the robot (and other active devices like tablets in CoWriter). 
These modules can read values given by mutual models in order to compute decisions. 
In the example of the CoWriter activity, these modules are given by the system that learns and generate letters, but we can add a module that generates micro-behaviours, another that decides to switch to a new activity (e.g. drawing with the robot),\dots. 
The following subsections explain in detail the content and operation of each part of the architecture. 

\subsection{Perception modules}
\label{ssec:perception}
The sensitive modules measure values of relevant perceived variables. 
While the agent's gaze direction and facial expression can be used in any interaction, some additive variables can be specific to the activity: in CoWriter, a module takes as input perceived variables from a tablet to compute the new state of robot's writing. 
It defines a sensitive module, and the value of the new state of the robot to write a letter defines a perceived variable. 
The evaluation of the robot by the child via the feedback buttons on the tablet defines another perceived variable provided by the modules of the activity. 
Other modules are independent of the activity: the system that estimates the target objects looked at by the child provides additive information not directly used by the modules of the activity.  

\subsection{Mutual modelling modules}
\label{ssec:mmm}
Each perceived value measured by the sensitive modules are associated with the model of an agent (or a $n^{th}$-order agent). Each mutual model can be designed as a module that deals with a list of associated perceived variables and watch if the value of one of these variables has been changed. 
An additive module knows all the expected causalities and computes the values of abstract variables. Some expected causalities can be empirically learned and others pre-programmed. 

\subsection{Decision making}
\label{ssec:decision}
The values of mutual modelling variables will provide rich and useful information for decision making. 
Taking in account these values to elaborate decision should improve the realism and the efficiency of the robot in the interaction.
Similarly to the sensitive ones, the modules that make decisions can be specific to the activity (in CoWriter, the choice of a new learning curve or the decision to suddenly make a big mistake), or can govern a general behaviour (for example the exaggeration of a misunderstood gesture). Some decisions can have a high impact on the interaction: to stop an activity and to switch to a new one can frustrate a child that was committed. 
The conditions to make such a decision are not directly assessable, but must be learned by the robot. In order to make these decision cautiously, we propose to start by a Wizard-of-Oz approach and to move towards an autonomous approach following these steps: 
\begin{enumerate}
\item \textbf{Wizard-of-Oz}: A human takes decisions; the robot learns
\item \textbf{Mixed-initiative}: The robot makes suggestions; a human agrees or disagrees
\item \textbf{Autonomous}: The robot makes decisions
\end{enumerate}

\section{Conclusion}

Educational HRI based on learning by teaching approach needs robots to be able to perform second-level mutual modelling. 
We introduced a new approach to implement mutual modelling into a cognitive architecture. We used the CoWriter activity
as an example of application, but our architecture could be easily generalised for any kind of interaction.
We believe that this step must be reached in other contexts of HRI, in order to develop higher realism of behaviours
and to improve the quality of interactions.

\section*{Acknowledgment}
This research was partially supported by the Funda\c{c}\~{a}o para a Ci\^{e}ncia
e a Tecnologia (FCT) with reference UID/CEC/ 50021/2013, and by the Swiss
National Science Foundation through the National Centre of Competence in
Research Robotics.

\bibliographystyle{IEEEtran}

\bibliography{biblio}

\begin{thebibliography}{10}
\providecommand{\url}[1]{#1}
\csname url@samestyle\endcsname
\providecommand{\newblock}{\relax}
\providecommand{\bibinfo}[2]{#2}
\providecommand{\BIBentrySTDinterwordspacing}{\spaceskip=0pt\relax}
\providecommand{\BIBentryALTinterwordstretchfactor}{4}
\providecommand{\BIBentryALTinterwordspacing}{\spaceskip=\fontdimen2\font plus
\BIBentryALTinterwordstretchfactor\fontdimen3\font minus
  \fontdimen4\font\relax}
\providecommand{\BIBforeignlanguage}[2]{{%
\expandafter\ifx\csname l@#1\endcsname\relax
\typeout{** WARNING: IEEEtran.bst: No hyphenation pattern has been}%
\typeout{** loaded for the language `#1'. Using the pattern for}%
\typeout{** the default language instead.}%
\else
\language=\csname l@#1\endcsname
\fi
#2}}
\providecommand{\BIBdecl}{\relax}
\BIBdecl

\bibitem{baron1985does}
S.~Baron-Cohen, A.~Leslie, and U.~Frith, ``Does the autistic child have a
  ``theory of mind'' ?'' \emph{Cognition}, 1985.

\bibitem{dillenbourg1999you}
P.~Dillenbourg, ``What do you mean by collaborative learning?''
  \emph{Collaborative-learning: Cognitive and Computational Approaches.}, pp.
  1--19, 1999.

\bibitem{Hood}
D.~Hood, S.~Lemaignan, and P.~Dillenbourg, ``When children teach a robot to
  write: An autonomous teachable humanoid which uses simulated handwriting,''
  in \emph{Proceedings of the Tenth Annual ACM/IEEE International Conference on
  Human-Robot Interaction}, ser. HRI '15.\hskip 1em plus 0.5em minus
  0.4em\relax New York, NY, USA: ACM, 2015, pp. 83--90.

\bibitem{jacq2016building}
A.~Jacq, S.~Lemaignan, F.~Garcia, P.~Dillenbourg, and A.~Paiva, ``Building
  successful long child-robot interactions in a learning context,'' in
  \emph{Proceedings of the 2016 ACM/IEEE Human-Robot Interaction Conference},
  2016.

\bibitem{Rohrbeck2003}
\BIBentryALTinterwordspacing
C.~A. Rohrbeck, M.~D. Ginsburg-Block, J.~W. Fantuzzo, and T.~R. Miller,
  ``{Peer-assisted learning interventions with elementary school students: A
  meta-analytic review},'' \emph{Journal of Educational Psychology}, vol.~95,
  no.~2, pp. 240--257, 2003. [Online]. Available:
  \url{http://doi.apa.org/getdoi.cfm?doi=10.1037/0022-0663.95.2.240}
\BIBentrySTDinterwordspacing

\bibitem{lemaignan2015mutual}
S.~Lemaignan and P.~Dillenbourg, ``Mutual modelling in robotics: Inspirations
  for the next steps,'' in \emph{Proceedings of the 2015 ACM/IEEE Human-Robot
  Interaction Conference}, 2015.

\bibitem{flavell1990developmental}
J.~H. Flavell, F.~L. Green, and E.~R. Flavell, ``Developmental changes in young
  children's knowledge about the mind,'' \emph{Cognitive Development}, vol.~5,
  no.~1, pp. 1--27, 1990.

\bibitem{hendricks2008epistemic}
V.~Hendricks and J.~Symons, ``Epistemic logic,'' in \emph{Stanford Encyclopedia
  of Philosophy}, 2008.

\bibitem{roschelle1995construction}
J.~Roschelle and S.~D. Teasley, ``The construction of shared knowledge in
  collaborative problem solving,'' in \emph{Computer supported collaborative
  learning}.\hskip 1em plus 0.5em minus 0.4em\relax Springer, 1995, pp. 69--97.

\bibitem{sangin2007partner}
M.~Sangin, N.~Nova, G.~Molinari, and P.~Dillenbourg, ``Partner modeling is
  mutual,'' in \emph{Proceedings of the 8th iternational conference on
  {C}omputer {S}upported {C}ollaborative {L}earning}.\hskip 1em plus 0.5em
  minus 0.4em\relax International Society of the Learning Sciences, 2007, pp.
  625--632.

\bibitem{scassellati2002theory}
B.~Scassellati, ``Theory of mind for a humanoid robot,'' \emph{Autonomous
  Robots}, vol.~12, no.~1, pp. 13--24, 2002.

\bibitem{breazeal2006using}
C.~Breazeal, M.~Berlin, A.~Brooks, J.~Gray, and A.~Thomaz, ``Using perspective
  taking to learn from ambiguous demonstrations,'' \emph{Robotics and
  Autonomous Systems}, pp. 385--393, 2006.

\bibitem{Trafton2005}
J.~Trafton, N.~Cassimatis, M.~Bugajska, D.~Brock, F.~Mintz, and A.~Schultz,
  ``Enabling effective human-robot interaction using perspective-taking in
  robots,'' \emph{IEEE Transactions on Systems, Man and Cybernetics, Part A:
  Systems and Humans}, vol.~35, no.~4, pp. 460--470, 2005.

\bibitem{Ros2010}
R.~Ros, E.~A. Sisbot, R.~Alami, J.~Steinwender, K.~Hamann, and F.~Warneken,
  ``Solving ambiguities with perspective taking,'' in \emph{5th ACM/IEEE
  International Conference on Human-Robot Interaction}, 2010.

\bibitem{lemaignan2012thesis}
S.~Lemaignan, ``Grounding the interaction: Knowledge management for interactive
  robots,'' Ph.D. dissertation, {CNRS} - Laboratoire d'Analyse et
  d'Architecture des Systèmes, Technische Universität München - Intelligent
  Autonomous Systems lab, 2012.

\bibitem{flavell1977development}
J.~H. Flavell, ``\BIBforeignlanguage{eng}{The development of knowledge about
  visual perception.}'' \emph{\BIBforeignlanguage{eng}{Nebraska Symposium on
  Motivation}}, vol.~25, pp. 43--76, 1977.

\bibitem{breazeal2009embodied}
C.~Breazeal, J.~Gray, and M.~Berlin, ``An embodied cognition approach to
  mindreading skills for socially intelligent robots,'' \emph{The International
  Journal of Robotics Research}, vol.~28, no.~5, pp. 656--680, 2009.

\bibitem{warnier2012when}
M.~Warnier, J.~Guitton, S.~Lemaignan, and R.~Alami, ``When the robot puts
  itself in your shoes. managing and exploiting human and robot beliefs,'' in
  \emph{Proceedings of the 21st IEEE International Symposium on Robot and Human
  Interactive Communication}, 2012, pp. 948--954.

\bibitem{wimmer1983beliefs}
H.~Wimmer and J.~Perner, ``Beliefs about beliefs: Representation and
  constraining function of wrong beliefs in young children's understanding of
  deception,'' \emph{Cognition}, vol.~13, no.~1, pp. 103--128, 1983.

\bibitem{lemaignan2016realtime}
S.~Lemaignan, F.~Garcia, A.~Jacq, and P.~Dillenbourg, ``From real-time
  attention assessment to “with-me-ness” in human-robot interaction,'' in
  \emph{Proceedings of the 2016 ACM/IEEE Human-Robot Interaction Conference},
  2016.

\bibitem{sharma2014me}
K.~Sharma, P.~Jermann, and P.~Dillenbourg, ``“with-me-ness”: A gaze-measure
  for students’ attention in moocs,'' in \emph{International conference of
  the learning sciences}, no. EPFL-CONF-201918, 2014.

\bibitem{clark1991grounding}
H.~H. Clark and S.~E. Brennan, ``Grounding in communication,''
  \emph{Perspectives on socially shared cognition}, vol.~13, no. 1991, pp.
  127--149, 1991.

\end{thebibliography}

\end{document}